\documentclass[letterpaper, 10 pt, conference]{ieeeconf}  

\IEEEoverridecommandlockouts                              

\usepackage{graphicx} 
\usepackage{amsmath} 
\usepackage{amssymb}  
\usepackage{subcaption}
\usepackage{siunitx}
\usepackage{cite}
\usepackage{bm}
\usepackage{mathtools}
\usepackage[font=footnotesize]{caption}
\usepackage[abbreviations]{glossaries-extra}
\usepackage{flushend}
\usepackage{booktabs}
\usepackage[colorlinks = true,
            linkcolor = blue,
            urlcolor  = blue,
            citecolor = blue,
            anchorcolor = blue]{hyperref}
\usepackage[capitalise]{cleveref}

\title{Pedipulate: Enabling Manipulation Skills using a Quadruped Robot's Leg}

\author{Philip Arm, Mayank Mittal, Hendrik Kolvenbach, and Marco Hutter
\thanks{All authors are with ETH Zurich, Robotics Systems Lab; Leonhardstrasse 21, 8092 Zurich, Switzerland. M. Mittal is also with NVIDIA.
        Contact: {\tt\small parm@ethz.ch}}%
\thanks{This research was supported by the Swiss National Science Foundation through the National Centre of Competence in Digital Fabrication (NCCR dfab). This project has received funding through ESA contract nos. 4000137333/22/NL/AT and 4000135310/21/NL/PA/pt. This work has been conducted as part of ANYmal Research, a community to advance legged robotics.
}
}
\begin{document}

\bstctlcite{IEEEexample:BSTcontrol}

\maketitle
\thispagestyle{empty}
\pagestyle{empty}

\begin{abstract}
Legged robots have the potential to become vital in maintenance, home support, and exploration scenarios. In order to interact with and manipulate their environments, most legged robots are equipped with a dedicated robot arm, which means additional mass and mechanical complexity compared to standard legged robots. In this work, we explore pedipulation - using the legs of a legged robot for manipulation. By training a reinforcement learning policy that tracks position targets for one foot, we enable a dedicated pedipulation controller that is robust to disturbances, has a large workspace through whole-body behaviors, and can reach far-away targets with gait emergence, enabling loco-pedipulation. By deploying our controller on a quadrupedal robot using teleoperation, we demonstrate various real-world tasks such as door opening, sample collection, and pushing obstacles. We demonstrate load carrying of more than \SI{2.0}{\kilo \gram} at the foot. Additionally, the controller is robust to interaction forces at the foot, disturbances at the base, and slippery contact surfaces. Videos of the experiments are available at \href{https://sites.google.com/leggedrobotics.com/pedipulate}{https://sites.google.com/leggedrobotics.com/pedipulate}.

\end{abstract}

\section{INTRODUCTION}
In recent years, the locomotion capabilities of legged robots have greatly improved. Specifically, quadrupedal robots are nowadays capable of traversing various industrial and natural environments~\cite{lee2020learning, miki2022learning, choi2023learning, hoeller2023anymal}, facilitating their deployment in inspection, exploration, and search-and-rescue missions~\cite{hutter2018towards, kolvenbach2020jfr, stachowiak2021procedures, tranzatto2022cerberus, arm2023scientific, lindqvist2022multimodality}. Although legged robots have reached industrial maturity, their application range currently remains limited to inspection tasks that require minimal interaction with the environment. We need to improve and robustify their manipulation capabilities to deploy legged robots in a broader range of real-world scenarios, such as maintenance, home support, and sample collection.

Recent works in legged mobile manipulation have mainly focused on systems with a dedicated robotic arm for interactive tasks such as door opening, fetching objects, and opening valves~\cite{bellicoso2019alma, zimmermann2021go, mittal2022articulated, ferrolho2022roloma, sleiman2023versatile}. However, such an additional arm increases the robot's mechanical complexity and power consumption. Taking inspiration from quadrupedal animals~\cite{pets}, we hypothesize that many manipulation tasks do not require the additional dexterity and complexity of robotic arms and hands but can be solved using the legs of quadrupedal robots. Using the same limbs for locomotion and manipulation can reduce a robotic system's mechanical complexity and cost, which is especially beneficial in mass-constrained applications such as space exploration.
\begin{figure}[t]
    \centering
    \includegraphics[width=0.48\textwidth]{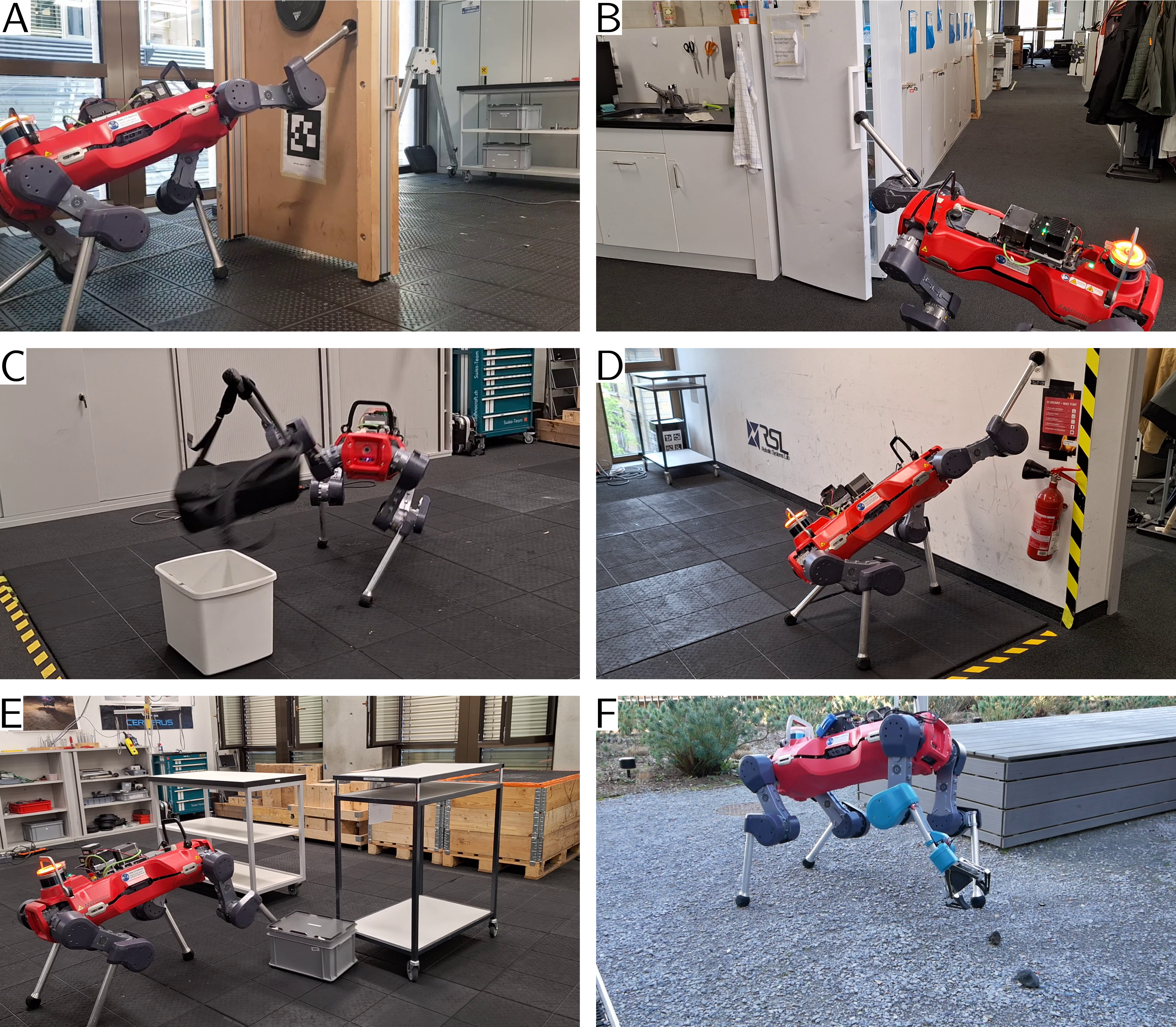}
    \caption{Our foot target tracking controller enables a variety of real-world manipulation tasks such as opening doors (A) and fridges (B), object transport (C), pressing a button (D), pushing obstacles out of the way (E), and collecting rock samples (F).}
    \label{fig:eyecatcher}
\end{figure}
\begin{figure*}[t]
    \centering
    \includegraphics[width=\textwidth]{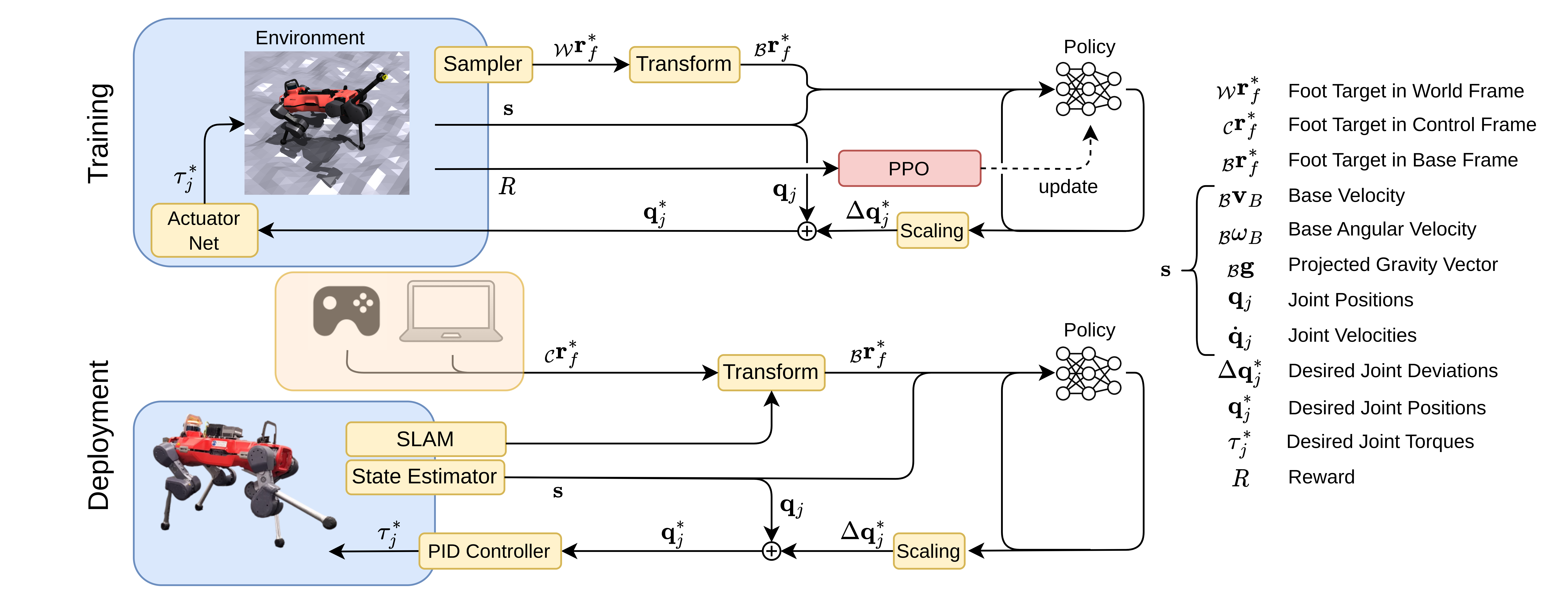}
    \caption{Overview of our training and deployment setup. We first specify the commands in an inertial frame, namely the world frame in training and the control frame during deployment. To make the policy agnostic to the used inertial frame, we transform the commands to the base frame before adding them to the observations. The policy's actions are interpreted as deviations from the current joint position.}
    \label{fig:pipeline}
\end{figure*}

We, therefore, seek to enable a broad range of manipulation tasks using the legs of quadrupedal robots - so-called \textit{pedipulation}. By developing a versatile low-level controller that can track foot target points, we investigate how we can use pedipulation to enable real-world skills. To this end, we use a deep reinforcement learning setup to train a robust controller for foot target point tracking. We additionally seek to enable the controller to approach far-away targets with the foot in the air while using a tripod gait, facilitating object transport. We investigate the controller's tracking performance and disturbance rejection capabilities in simulation and hardware experiments and deploy it in various real-world tasks through teleoperation. Specifically, our key contributions are as follows:

\begin{itemize}
    \item We design a reinforcement learning-based pedipulation controller by training a policy that tracks foot position targets.
    \item We investigate the achievable workspace of our controller while standing and show that it reaches a large local workspace through inherent whole-body motions.
    \item We enable the controller to adapt the robot's stance and locomote towards far-away targets via curriculum-based command sampling.
    \item We demonstrate that the controller is robust to external disturbances, such as interaction forces on the base and foot, and slippery terrains.
    \item We show that through the above features, our learned pedipulation controller enables numerous real-world manipulation skills such as door opening, rock sample collection, and pushing obstacles without task-specific adaptions~(\cref{fig:eyecatcher}).

\end{itemize}

\section{RELATED WORK}
Several works investigated manipulation using robotic legs - or, more precisely, pedipulation - in recent years. We split these works into three categories:
\begin{itemize}
    \item \textit{Manipulation via locomotion}: The robot moves objects towards a goal by pushing them with its body while walking.
    \item \textit{Non-prehensile pedipulation}: The robot manipulates objects with its legs without the capability to exert pulling forces onto the objects.
    \item \textit{Prehensile pedipulation}: The robot has a specialized tool on one or multiple legs to grasp and manipulate objects with pushing and pulling forces.
\end{itemize}

\subsection{Manipulation via Locomotion}
Manipulation via locomotion can enable robots to push objects close to their body weight~\cite{jeon2023learning}. Scaling this strategy to a multi-robot approach allows manipulating even larger objects~\cite{nachum2019multi, mataric1995cooperative}. In these works, the authors trained a reinforcement learning policy to move large-scale objects to a desired 2D pose. This approach can, for example, be used to remove obstacles. However, it does not generalize to dexterous tasks such as opening doors or pushing small buttons, which we aim to tackle in this paper.

\subsection{Non-prehensile Pedipulation}
Non-prehensile pedipulation facilitates moving obstacles out of the way~\cite{lu2020autonomous}, kicking or balancing balls~\cite{ji2022hierarchical, shi2021circus}, probing the environment~\cite{kolvenbach2019haptic, kolvenbach2020jfr}, or opening doors~\cite{topping2017quasi}. Some of the above approaches track pre-defined trajectories with a tracking controller. Lu et al.~\cite{lu2020autonomous} used an inverse kinematics controller to track a heuristically defined Bézier Curve to push objects on the ground. Kolvenbach et al.~\cite{kolvenbach2019haptic, kolvenbach2020jfr} executed a pre-defined trajectory with FreeGait, an API for whole-body control~\cite{freeGait} to haptically inspect the terrain. Others trained a reinforcement learning policy on a single task, for example, kicking balls~\cite{ji2022hierarchical} or balancing balls~\cite{shi2021circus}. While many of these works show promising results in their respective domain, our goal is to have a general controller that can be used for multiple manipulation tasks.

Cheng et al.~\cite{cheng2023legs} showed a non-prehensile pedipulation controller capable of generalizing to multiple tasks. In this work, the authors used separately learned low-level manipulation and locomotion policies and a high-level behavior tree to synthesize skills such as kicking a ball or pushing buttons. While the paper shows an impressive range of skills, the low-level manipulation and locomotion policies are separated, which implies that combined locomotion and pedipulation maneuvers (loco-pedipulation) are not feasible. Furthermore, the manipulation policy only tracks half-sine trajectories near the robot's nominal foot position. This form of command sampling probably does not generalize to arbitrary trajectories in the robot's workspace. Both these limitations prevent the application of the approach in more complex scenarios, as considered in this work, like door opening or object transport.

\subsection{Prehensile Pedipulation}
Integrating a gripper on the robot's foot enables applications such as fetching objects or collecting samples~\cite{roennau2014lauron, heppner2014versatile, brinkmann2020enhancement, tsvetkov2022novel}. Both the Lauron robot~\cite{roennau2014lauron, heppner2014versatile} and Mantis~\cite{brinkmann2020enhancement} are hexapod robots where the integrated grippers can fetch objects. These works show capable gripper concepts. However, they provide little detail on their manipulation control approach and do not demonstrate robustness to external disturbances. Tsvetkov and Ramamoorthy~\cite{tsvetkov2022novel} introduced a dactylus-style gripper for single-leg and two-leg manipulation. In this work, the gripper concept is remarkable and facilitates manipulating multiple objects and tools. However, the operator can only command the robot's target position in joint space. The target position is converted into a trajectory and tracked in an open-loop fashion. This approach requires cumbersome command tuning and limits the robustness to disturbances. In our work, we aim to overcome these issues and develop a versatile controller which is useful for both prehensile and non-prehensile pedipulation scenarios.

\section{METHOD}
We hypothesize that a hierarchical structure with a task-specific high-level planner and a generic low-level controller generalizes well to numerous pedipulation tasks. Here, we focus on the required low-level controller that effectively tracks foot position commands and is robust to disturbances. Due to the recent success of reinforcement learning~\cite{lee2020learning, miki2022learning, hoeller2023anymal, agarwal2023legged, choi2023learning}, we choose to use deep reinforcement learning to train a neural network policy (\cref{fig:pipeline}). We train the policy to track foot target points of the right front foot with minimal additional rewards. During deployment, the user provides the command in an inertial frame, and we expect high-level planners to do the same. However, we want the policy to be agnostic to the used inertial frame. Therefore, we internally transform the command to the robot's base frame before adding it to the observation vector. 
\\
We use Isaac Gym as a simulation environment~\cite{rudin2022learning, makoviychuk2021isaac} and PPO as a deep reinforcement learning algorithm~\cite{schulman2017proximal}. We use MLPs for the actor and critic networks as in previous work on locomotion~\cite{rudin2022learning} and rely on similar hyperparameters. We randomize simulation friction parameters, add random pushes to the robot's base in simulation, and train on irregular terrain to facilitate the sim-to-real transfer. We deployed our controller on the quadrupedal robot ANYmal D by ANYbotics~\cite{ANYmal}.

\begin{figure}[t]
    \centering
    \includegraphics[width=0.5\textwidth]{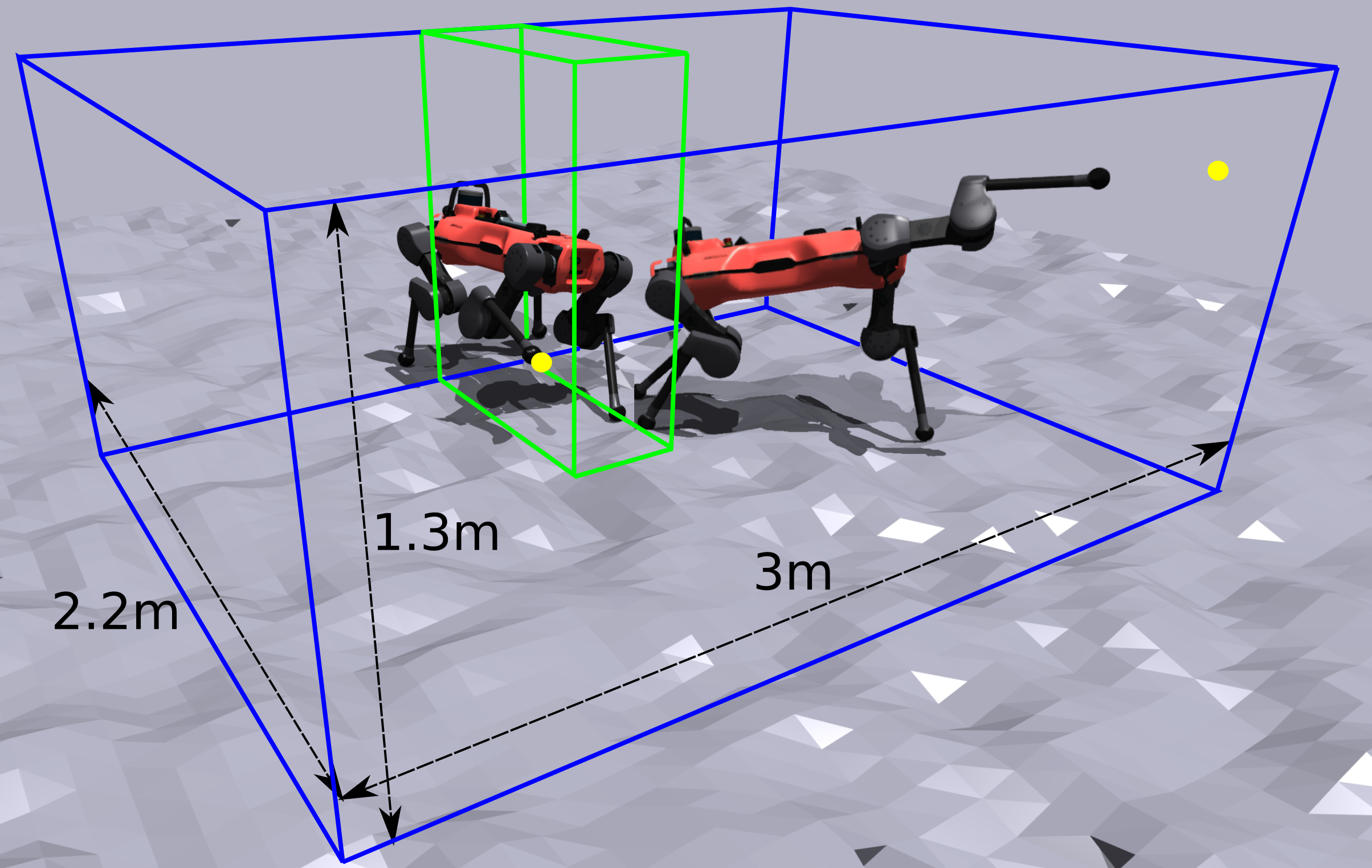}
    \caption{Simulation setup in Isaac Gym. The initial and final sampling space of the foot position commands are visualized in green and blue, respectively. The command curriculum enables combined pedipulation and tripod locomotion in a single policy. The figure shows one robot tracking a close target point (left) and a robot approaching a far-range target point using the tripod gait (right). The target points are visualized in yellow.}
    \label{fig:training_environment}
\end{figure}
\begin{figure*}[t]
    \centering
    \includegraphics[width=\textwidth]{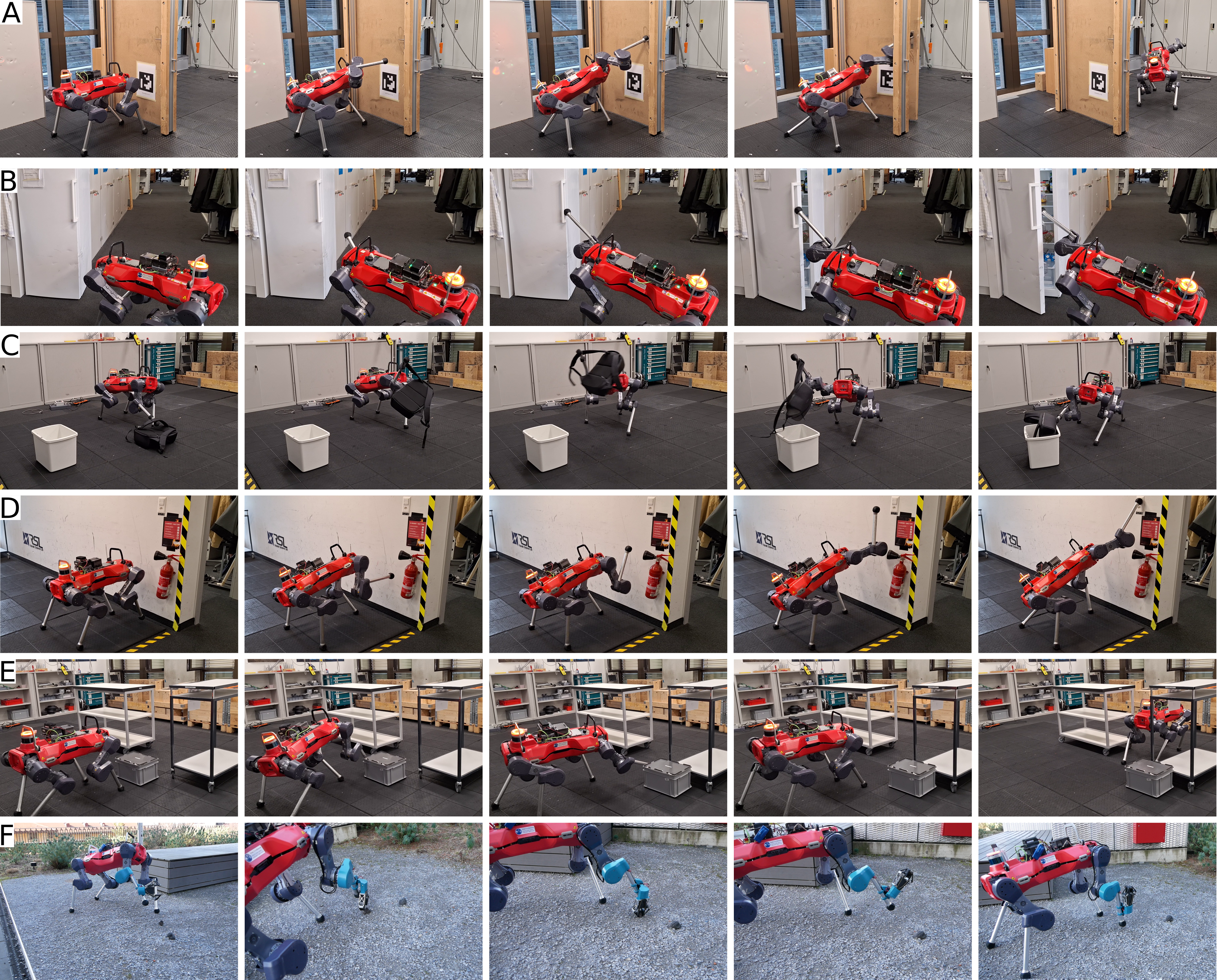}
    \caption{Our controller enables numerous real-world manipulation tasks: (A) The robot opens a push door. (B) The robot opens a fridge. (C) The robot lifts a backpack and transports it to a box using a tripod-hopping gait. (D) The large workspace of the controller allows pressing a button far above the robot's base. (E) The controller can be used to push obstacles out of the way. (F) With an additional gripper, the robot can collect rock samples.}
    \label{fig:task-overview}
\end{figure*}

\subsection{Observation and Action Space}
The policy receives proprioceptive information about the robot's state and the foot position commands as observations~(\cref{tab:obs_actions}). The scaled actions are interpreted as desired deviations from the current joint positions. We tested two action spaces: deviations from current joint positions and nominal joint positions. However, we did not notice a relevant difference in the achieved rewards and behaviors.

\begin{table}[h]
    \centering
    \begin{tabular}{c|c|c}
         \toprule
         \textbf{Observations} & & \\
         \midrule
         base linear velocity & $\prescript{}{\mathcal{B}}{\bm{v}_B}$ & $\mathbb{R}^3$ \\
         base angular velocity & $\prescript{}{\mathcal{B}}{\bm{\omega}_B}$ & $\mathbb{R}^3$ \\
         projected gravity vector & $\prescript{}{\mathcal{B}}{\bm{g}}$ & $\mathbb{R}^3$ \\
         joint positions & $\bm{q}_j$ & $\mathbb{R}^{12}$ \\
         joint velocities & $\dot{\bm{q}_j}$ & $\mathbb{R}^{12}$ \\
         foot position command & $\prescript{}{\mathcal{B}}{\bm{r}^*_f}$ & $\mathbb{R}^{3}$ \\
         last actions & $\bm{a}_{t-1}$ &$\mathbb{R}^{12}$ \\
         \midrule \midrule
        \textbf{Actions}& & \\
        \midrule
        desired deviation from last joint positions & $\Delta \bm{q}_j^*$&$\mathbb{R}^{12}$ \\
        \bottomrule
    \end{tabular}
    \caption{Observations and actions we use for our foot target tracking policy. The actor and critic receive the same set of observations.}
    \label{tab:obs_actions}
\end{table}

\subsection{Rewards}
We use two types of rewards; a task reward $R_e$ representing the main objective, and penalties $R_n$, which render the behavior safe and physically feasible on the real system. The total reward is $R = R_e + R_n$.

\subsubsection{Task Reward}
We provide a single dense task reward to encourage the foot to be close to the target point. The reward function is defined as 
\begin{align}
R_e = w_1 \cdot e^{-\left( \frac{  || \prescript{}{\mathcal{B}}{\bm{r}}_f-\prescript{}{\mathcal{B}}{\bm{r}}_f^* ||}{\sigma}\right)} ,
\end{align}
where $\prescript{}{\mathcal{B}}{\bm{r}}_f$ and $\prescript{}{\mathcal{B}}{\bm{r}}_f^*$ denote the measured and desired foot position in the base frame, respectively. We use $w_1 = 15$ and $\sigma$ = \SI{0.8}{\per \metre}.
\subsubsection{Normalization Rewards}
To reduce jerky motions and enable safe behaviors on the real robot, we add penalties on quantities such as joint torques, joint velocities, and collisions between robot links. The normalization reward is defined as
\begin{align}
\begin{split}
R_n &= w_2 ||\dot{\bm{q}_j}||^2 + w_3||\ddot{\bm{q}_j}||^2 + w_4 ||\bm{\tau}||^2 \\ + &w_5 ||\bm{a}_t - \bm{a}_{t-1}||^2  + w_6 \cdot n_c + w_7 \cdot k ,
\end{split}
\end{align}
where $\dot{\bm{q}_j}$ and $\ddot{\bm{q}_j}$ denote the joint velocities and accelerations, $\bm{\tau}$ denotes the joint torques, $\bm{a}_t$ the action at time $t$ and $w_i$ the scaling factors. Additionally, we penalize collisions between shanks and thighs, expressed by the number of collisions at the current time step $n_c$. Lastly, we add a termination penalty, which is applied to the environments where a base collision is registered, denoted by $k \in \{0, 1\}$, where $k=1$ represents a termination at the current time step. We use the manually tuned reward scales
\begin{align*}
    w_2 &= -5 \cdot 10^{-2} \qquad  &&w_3 = -5 \cdot 10^{-6}\\
    w_4 &= -2 \cdot 10^{-5}   &&w_5 = -1 \cdot 10^{-2} \\
    w_6 &= -2  &&w_7 = -80 .
\end{align*}

\subsection{Terrain}
We observed that training on flat terrain leads to behaviors where stance feet slide across the ground to adapt the stance. On the real robot, such policies are prone to stumbling on terrain irregularities. To robustify the behavior, we train on irregular terrain~(\cref{fig:training_environment}). We uniformly sample a height field between \SI{-0.08}{\metre} and \SI{0.08}{\metre} on a grid with \SI{0.2}{\metre} resolution to create the terrain.

\subsection{Command Curriculum for Loco-Pedipulation}
To enable tracking far-away targets that require the robot to adapt the stance legs or walk on three legs, we implemented a curriculum for adaptive command sampling. Initially, commands are sampled uniformly in a box that is reachable for the foot without moving the stance feet~(green box in \cref{fig:training_environment}). Once the agent learns to track the commands with an average error of less than \SI{0.06}{\metre}, the command sampling box is extended by \SI{0.2}{m} in the robot's heading direction and \SI{0.2}{m} in both lateral directions. We limit the box to \SI{2.2}{m} x \SI{3.0}{m} x \SI{1.3}{m} (blue box in \cref{fig:training_environment}) and let the training converge with this command range. The command range does not limit the range of loco-pedipulation maneuvers since we can still give a sequence of target points to reach a further target.

\subsection{Sim-to-Real Transfer, Training, and Deployment}
We randomize the friction parameters in the simulation and add noise to the observations. Additionally, we exert random pushes onto the robot's base every \SI{3.0}{\second}, which we simulate by setting a random twist to the base link. Each twist component is sampled in $[-0.6, 0.6]$ \SI{}{\metre \per \second} or  $[-0.6, 0.6]$ \SI{}{\radian\per\second}. Furthermore, we add a disturbance force at the feet of up to \SI{12}{N} that is constant over an episode. Finally, we use an actuator network to simulate the actuator dynamics as in~\cite{hwangbo2019learning}. We train with an episode length of \SI{12}{\second} and resample commands every \SI{6}{\second}. Training takes \SI{5.5}{\hour} on an NVIDIA GeForce RTX 2080 Ti for 8000 iterations.

During deployment, we define the command in a fixed local control frame ($\prescript{}{\mathcal{C}}{\bm{r}_F^*}$). When our pedipulation controller starts, it locks the control frame at the current base position, and our SLAM solution~\cite{compSLAM} continuously updates the transforms between the map, the control, and the base frame. We define the command by the joystick or the GUI in the control frame and transform it into the base frame ($\prescript{}{\mathcal{B}}{\bm{r}_F^*}$) before adding it to the policy observations. The control frame allows the user to intuitively control the foot in a fixed frame rather than having the behavior influenced by the current base pose. The user can use the joystick to incrementally move the target point or a GUI interface to directly command a single far-away target point (\cref{fig:pipeline}).The policy and the PID controller run at \SI{50}{Hz} and \SI{400}{Hz}, respectively.

\section{RESULTS AND DISCUSSION}

\subsection{Real-World Manipulation Skills}
Tracking foot target points in a large workspace via whole-body motions allows for solving various manipulation tasks (\cref{fig:task-overview}). This section presents multiple manipulation skills we achieved on the real system using our designed pedipulation controller and teleoperation interface. The operator had a direct line of sight to the robot in all experiments.

\subsubsection{Door Opening}
Door opening requires a controller that is robust to interaction forces with the environment. Additionally, reaching the door handle with the robot's foot requires a large workspace. Lastly, passing the door when it is open requires a combined loco-pedipulation behavior. \\
The robot was able to open a push door with a resistive spring-damper element at the hinge (\cref{fig:task-overview}A). We used the joystick interface to push down the handle and then set a target point behind the door using the GUI to move the robot through the door using the tripod gait while simultaneously pushing the door open with the pedipulating foot. \\
Similarly, we used our controller to open a fridge door (\cref{fig:task-overview}B). The friction on the foot was sufficient to open the pull door on the fridge. The interaction force at the door did not destabilize the controller.

\subsubsection{Transporting a Backpack}
In this experiment, we collected a \SI{2.0}{\kilo \gram} backpack by sliding the foot into one of the shoulder straps and raising the foot using the joystick interface (\cref{fig:task-overview}C). We then commanded a single target point above a box using the GUI interface and lowered the foot to release the backpack into the box with the joystick. This experiment demonstrates the ability of our controller to transport loads with the foot in the air thanks to the combined loco-pedipulation capability.

\subsubsection{Pressing a Hard-to-Reach Button}
The robot can exploit the large workspace of our pedipulation controller to reach and press a button of \SI{0.03}{\metre} x \SI{0.05}{\metre} at a height of \SI{1.25}{\metre} (\cref{fig:task-overview}D). The operator can move the target point into the wall such that the foot exerts a small force on the button without destabilizing the controller.


\subsubsection{Moving an Obstacle}
A common problem in legged locomotion is moving obstacles out of the way in confined spaces (\cref{fig:task-overview}C). Our pedipulation controller allows the user to move obstacles out of the robot's way to continue walking through a confined space. Again, robustness to unmodelled forces is required. In this experiment, we used a box with a mass of \SI{2.6}{\kilo \gram}.

\subsubsection{Rock Sample Collection}
By adding a gripper to the foot, the robot can conduct prehensile pedipulation tasks like collecting rock samples (\cref{fig:task-overview}D). This task is especially relevant in scenarios like planetary exploration, which require capable robots with minimal mass and mechanical complexity. We successfully collected rock samples of a size of \SI{0.08}{\metre} and a mass of roughly \SI{300}{\gram}. The controller proved robust to the additional mass of the gripper and the rock. 

\subsection{Reachable Workspace and Tracking Performance}
\begin{figure}[t]
    \centering
    \includegraphics[width=0.5\textwidth]{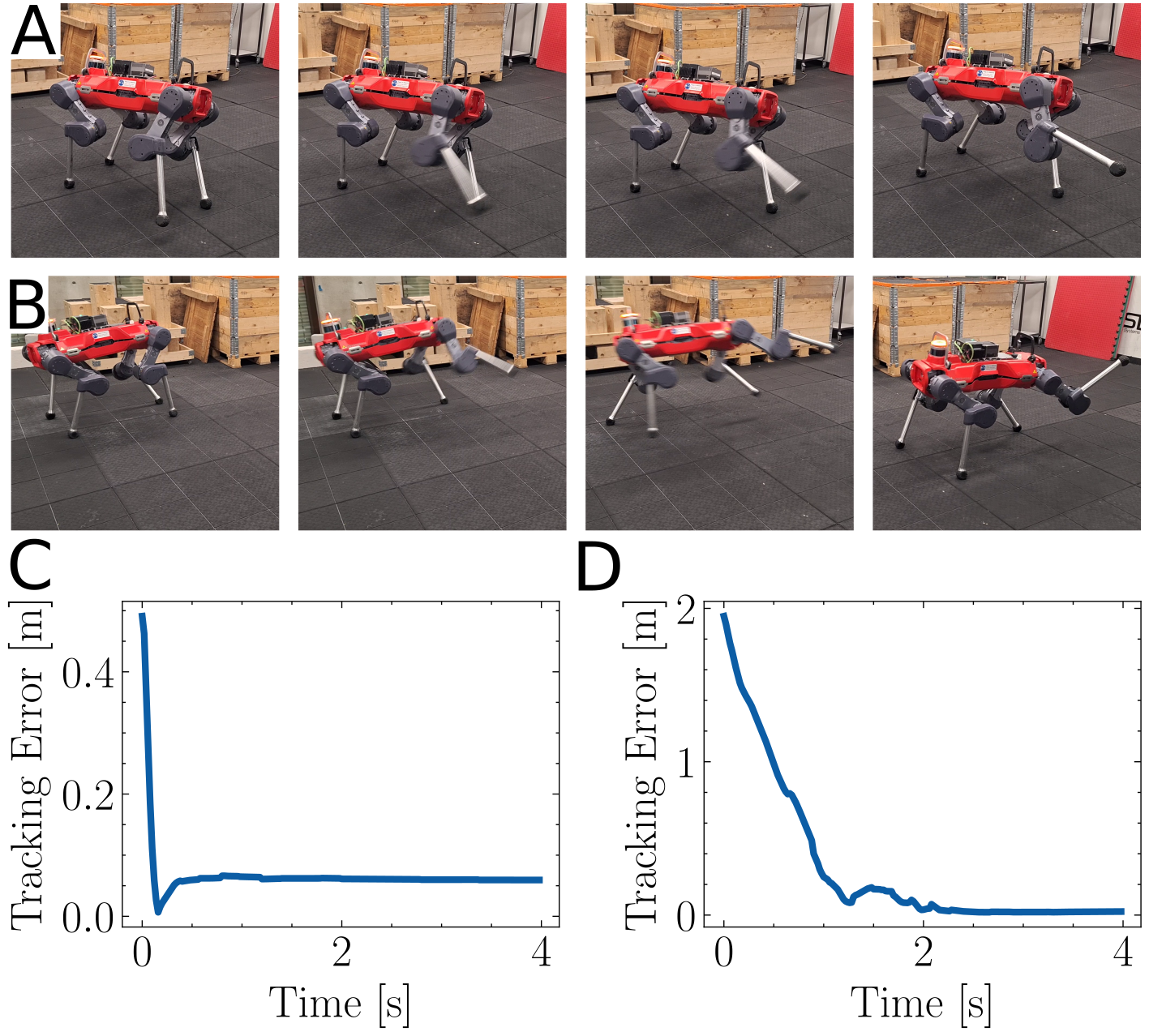}
    \caption{For close-range targets, the policy does not require a stance adaptation (A) and the tracking error quickly converges to a steady state value (C). For far-range targets, a tripod gait emerges (B), and the tracking error oscillates in the slow-down maneuver before converging (D).}
    \label{fig:step_response}
\end{figure}

We investigated the reachable workspace of our pedipulation controller and benchmarked the tracking performance. In these experiments, we distinguish between two behaviors: Close-range tracking and far-range tracking. Close-range tracking refers to target points the robot can reach without adapting its stance or walking to reach it. Far-range tracking refers to target points the policy reaches via locomotion or stance adaptation. \cref{fig:step_response} shows the respective behaviors. Notably, the transition between the two is seamless. If the target point moves outside the reachable range, the robot automatically starts stepping to follow the target point.

\subsubsection{Close-Range Target Tracking}
\begin{figure}[t]
    \centering
    \includegraphics[width=0.5\textwidth]{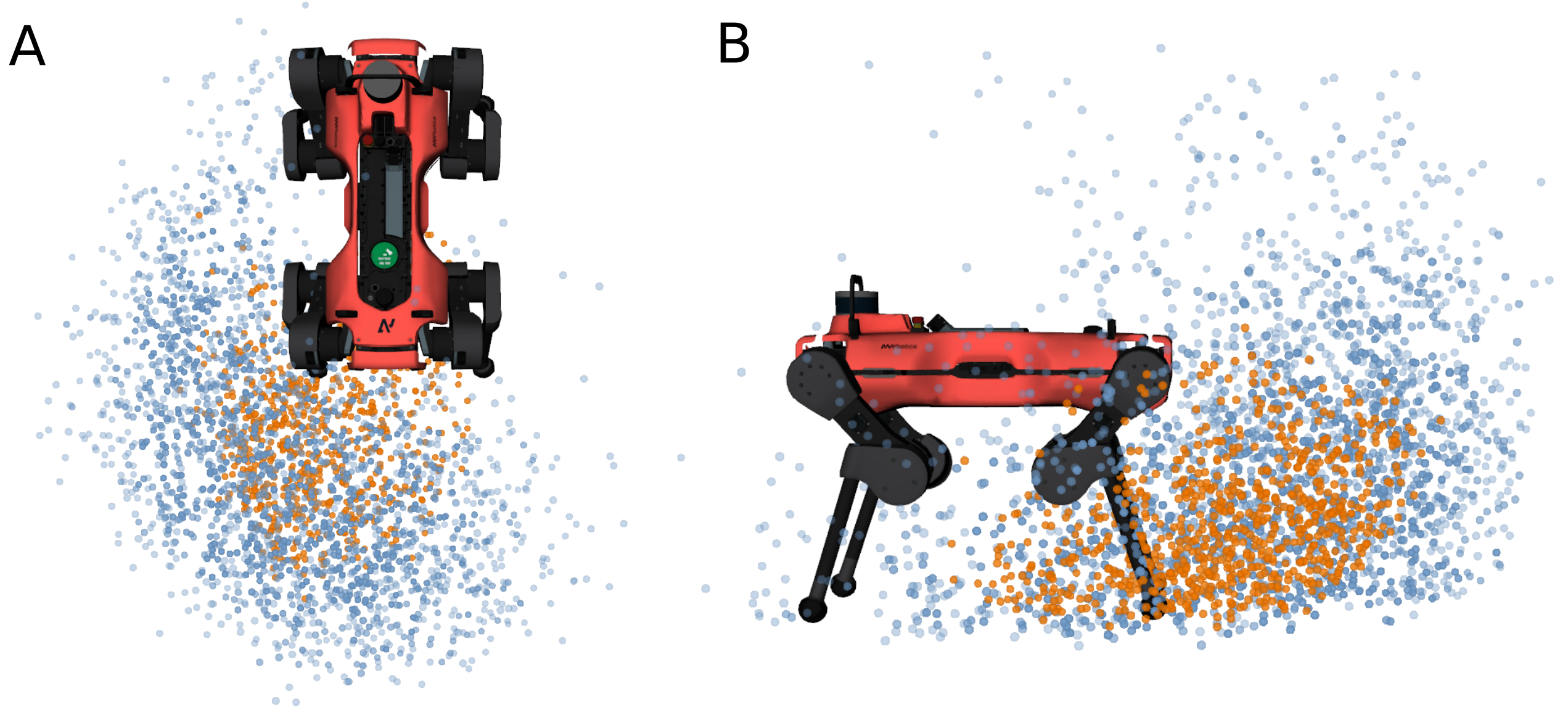}
    \caption{Thanks to the whole-body behavior, the robot reaches a large workspace even without the need to adapt its stance. The blue dots represent sampled commands the controller tracks without changing the stance configuration and with a tracking error below \SI{0.05}{\metre}, shown in top view (A) and side view (B). For reference, the orange dots are reachable samples for a fixed-base inverse kinematics controller with the same sampling space.}
    \label{fig:close_range_workspace}
\end{figure}

To evaluate the workspace our controller covers without locomotion, we deployed it in Isaac Gym and uniformly sampled commands in the robot's vicinity. After each episode, we discarded targets if the robot broke contact with one of the designated stance legs or if the tracking error was higher than \SI{0.05}{\metre}. As shown in \cref{fig:close_range_workspace}, the whole-body behavior of the controller allows the robot to reach a large workspace, even without adapting its stance. Note that this workspace does not coincide with the theoretical limits reachable through whole-body motions in stance. The robot stretches almost to the kinematic limits but starts stepping early to reach a more favorable configuration. These samples are discarded in this evaluation.

Additionally, we evaluated the tracking performance in the close-range workspace by sampling points within a box of \SI{0.3}{\metre} x \SI{0.4}{\metre} x \SI{0.2}{\metre} both in simulation and on the real robot. We compared the foot target position with the ground truth position in the simulation and the estimated position from the state estimator on the real robot. In simulation, the controller reached an average error of \SI{0.037}{\metre}, while on the real robot, the average error was \SI{0.057}{\metre}. We use a fixed-base inverse kinematics controller as a baseline, which reached an average error of \SI{0.041}{\metre} in simulation on the reachable range.

\subsubsection{Far-Range Target Tracking}
To evaluate the tracking performance of the controller on target points the policy reaches by stepping or walking, we evaluated the tracking error on the full training range (\SI{2.2}{m} x \SI{2.7}{m} x \SI{1.3}{m}) in Isaac Gym. We uniformly sampled commands on this range and logged the tracking error at the end of each episode. Our controller reaches an average tracking error of \SI{0.043}{\metre}.

\subsection{Robustness against Disturbances}
\begin{figure}[t]
    \centering
    \includegraphics[width=0.45\textwidth]{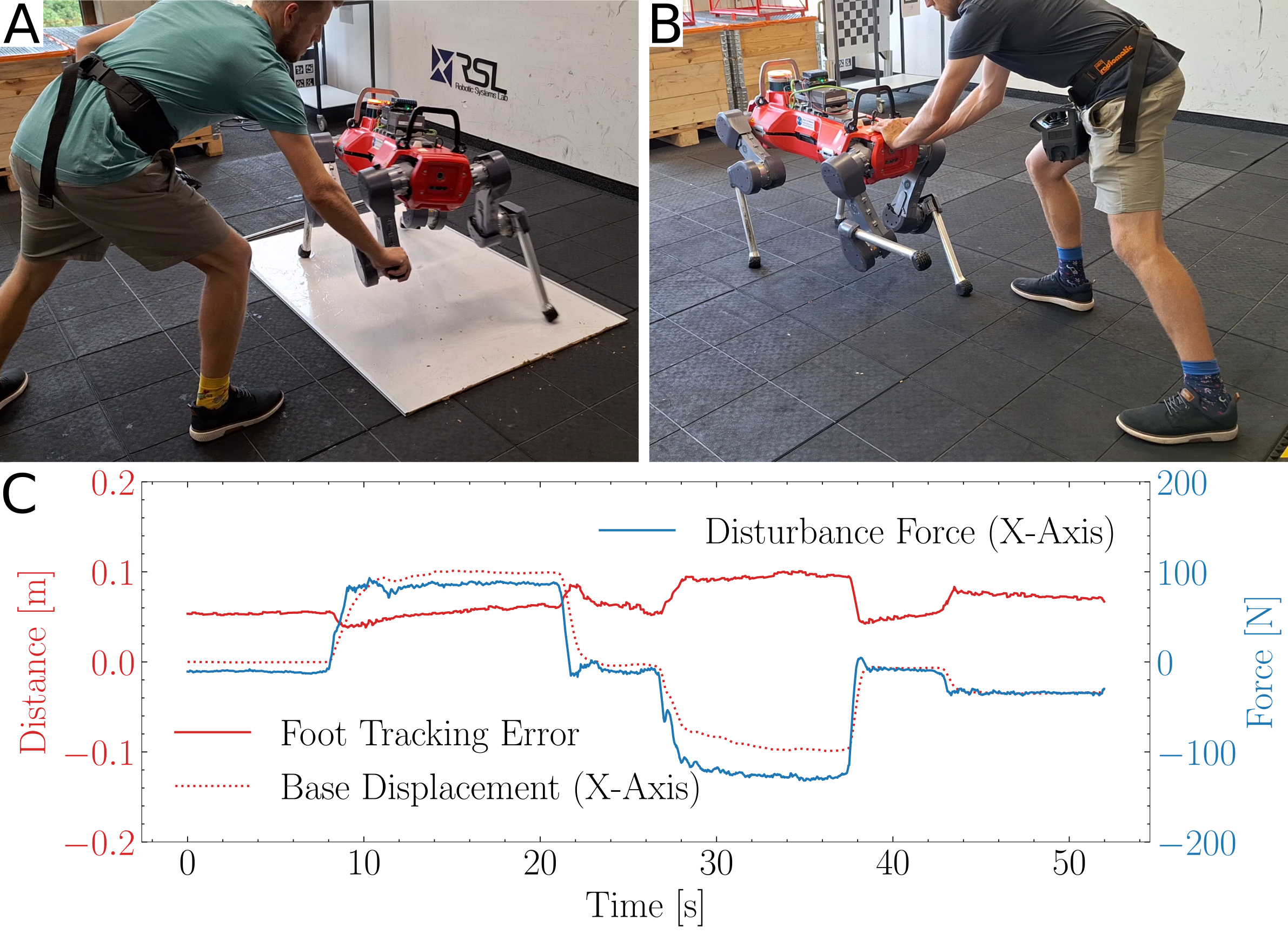}
    \caption{(A): The controller can react to disturbances, even on slippery surfaces like whiteboards. (B): When the robot's base is disturbed, the controller compensates for the base motion. (C): Evolution of the tracking error over time when the base is disturbed.}
    \label{fig:disturbance_rejection}
\end{figure}

To evaluate the controller's ability to generalize to different environmental conditions, we deployed it on ANYmal D standing on a wet whiteboard with extremely low friction. Although the stance legs slip during operation, the robot does not fall. Even when applying disturbance forces at the foot, which causes the stance legs to slip, the robot remains standing (\cref{fig:disturbance_rejection}A). \\
Additionally, we disturbed the robot's base while the controller was tracking a constant target point while standing on a flat floor. Thanks to the whole-body behavior of the controller, it can compensate for disturbances on the base. We applied disturbances of up to \SI{132}{N}. We estimated the force on the base by summing the contact forces on the stance legs we obtained from the state estimator via the joint torque measurements. The disturbance force resulted in a base displacement of up to \SI{0.1}{\metre} along the heading axis. Despite this disturbance, the tracking error only increased by \SI{0.06}{\metre} (\cref{fig:disturbance_rejection}C). Interestingly, the controller can compensate for disturbances in the negative heading direction better than for those in the positive heading direction. \\
If we disturb the base further, interesting behaviors emerge: If required, the stance feet adapt, and the robot steps to remain stable. The accompanying video shows this behavior.


\section{CONCLUSION AND FUTURE WORK}
In this work, we developed a deep reinforcement learning policy to track end-effector target points with a quadrupedal robot's foot. We showed that such a controller enables solving numerous real-world tasks via pedipulation: Using our controller, we could teleoperate the robot to successfully open doors, transport objects, press buttons, push obstacles, and collect rock samples. These skills are enabled by the controller's large workspace, achieved through inherent whole-body behaviors, its loco-pedipulation capability, and its robustness to disturbances. 
\\
While the controller is helpful in teleoperation scenarios, future work should move towards autonomous pedipulation. For such autonomy, a hierarchical approach, successfully used for other tasks~\cite{jeon2023learning}, could be a valid strategy. A hierarchical controller could contain our pedipulation policy as a low-level policy and combine it with task-specific perceptive high-level policies that output the foot targets. Additionally, we did not explicitly model interaction forces. Tracking interaction forces could extend the range of admissible manipulation tasks, such as haptic interaction or heavy payload transportation.
\\
Our work shows that numerous manipulation tasks can be solved by only doing pedipulation with quadrupedal robots. This insight will be relevant for future works on the design and control of legged mobile manipulators.

\section*{ACKNOWLEDGMENTS}
We thank Jonas Lauener for the mechanical design and control system design of the gripper and Ennio Schnieder for the design iteration on the gripper.

\bibliographystyle{IEEEtran}
\bibliography{references}
\end{document}